\documentclass{bmvc2k}

\usepackage{amssymb}
\usepackage{booktabs}
\usepackage{multirow}
\usepackage{float}
\usepackage{wrapfig} 

\definecolor{gray}{rgb}{0.5,0.5,0.5}
\definecolor{darkergreen}{RGB}{21, 152, 56}
\newcommand{\red}[1]{\textcolor{red}{#1}}
\newcommand{\gray}[1]{\textcolor{gray}{#1}}
\newcommand{\green}[1]{\textcolor[RGB]{96,177,87}{#1}}
\newcommand{\fn}[1]{\footnotesize{#1}}

\newcommand{\gbf}[1]{\green{\bf{\fn{(#1)}}}}
\newcommand{\rbf}[1]{\gray{\bf{\fn{(#1)}}}}
\newcommand{\redbf}[1]{\red{\bf{\fn{(#1)}}}}

\title{Exploring Localization for Self-supervised\\ Fine-grained Contrastive Learning}

\addauthor{Di Wu}{wudi@westlake.edu.cn}{1,2,$\star$}
\addauthor{Siyuan Li}{lisiyuan@westlake.edu.cn}{1,2,$\star$}
\addauthor{Zelin Zang}{zangzelin@westlake.edu.cn}{1,2}
\addauthor{Stan Z. Li}{Stan.ZQ.Li@westlake.edu.cn}{1,$\dag$}

\addinstitution{
AI Lab, School of Engineering,\\
Westlake University,\\
Hangzhou, Zhejiang, China
}
\addinstitution{
Zhejiang University,\\
Hangzhou, Zhejiang, China
}

\runninghead{Di Wu \textit{et al.}}{Cross-view Saliency Alignment}

\begin{document}

\maketitle

\begin{abstract}
Self-supervised contrastive learning has demonstrated great potential in learning visual representations. Despite their success in various downstream tasks such as image classification and object detection, self-supervised pre-training for fine-grained scenarios is not fully explored. We point out that current contrastive methods are prone to memorizing background/foreground texture and therefore have a limitation in localizing the foreground object. Analysis suggests that learning to extract discriminative texture information and localization are equally crucial for fine-grained self-supervised pre-training. Based on our findings, we introduce cross-view saliency alignment (CVSA), a contrastive learning framework that first crops and swaps saliency regions of images as a novel view generation and then guides the model to localize on foreground objects via a cross-view alignment loss. Extensive experiments on both small- and large-scale fine-grained classification benchmarks show that CVSA significantly improves the learned representation.
\let\thefootnote\relax\footnotetext{\hspace{-1.9em}$^\star$ The first two authors contribute equally\ \ \ \ $^\dag$ Corresponding author: Stan Z. Li}
\end{abstract}

\section{Introduction}
Learning visual representations without supervision by leveraging pretext tasks has become increasingly popular. Various learning approaches such as colorization \cite{eccv2016coloring}, Rel-Loc~\cite{noroozi2016unsupervised}, Rot-Pred~\cite{iclr2018rotation} have been proposed to learn such representations. The objective of these pretext tasks is to capture invariant features through predicting transformations applied to the same image. More recently, self-supervised representation learning has witnessed significant progress by the use of contrastive loss~\cite{hadsell2006dimensionality,logeswaran2018efficient,henaff2019data, cvpr2020moco, 2020simclr}. Despite that contrastive-based methods have even outperformed supervised methods under some circumstances, their success has largely been confined to large-scale general-purpose datasets (coarse-grained) such as ImageNet~\cite{krizhevsky2012imagenet}. We argue that current contrastive learning methods only work on coarse-grained iconic images with large foreground objects residing in the background with informative discriminative texture (\textit{e.g.,} ImageNet) but perform poorly when background texture provides little clue (\textit{e.g.,} CUB-200-2011~\cite{wah2011caltech}) for fine-grained separation.

\begin{figure}[t]
\centering
  \includegraphics[width=0.99\linewidth]{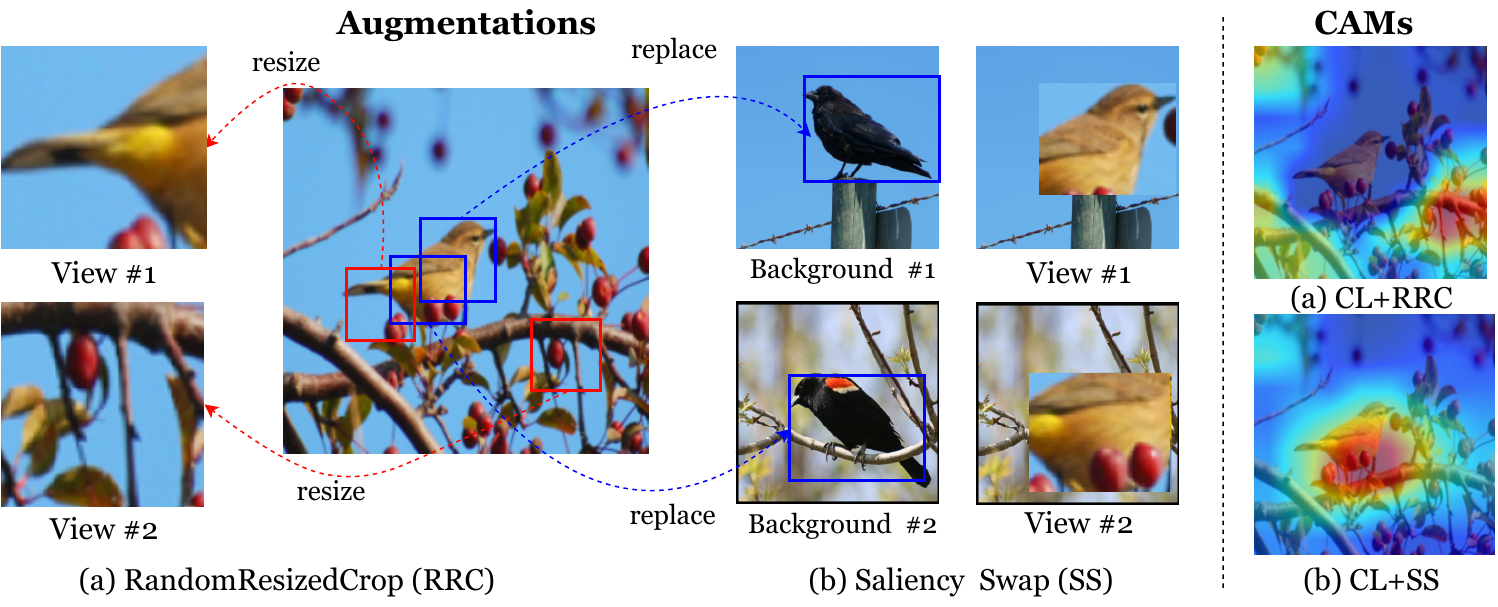}
  \vspace{-1em}
  \caption{
  Comparison of \textit{RandomResizedCrop} (RRC) and the proposed Saliency Swap (SS). We visualize Grad-CAM~\cite{iccv2017gradcam} of linear classifiers for pre-trained models.
  (a) shows the most commonly adopted RRC in contrastive learning (CL) methods. The random nature may cause views to contain mainly the backgrounds of the image leading to semantic inconsistency across views. (b) shows our proposed SS, which crops from regions of interest of the reference image and replaces the saliency regions of two randomly selected background images to guarantee semantic consistency.
  }
  \label{fig_1}
  \vspace{-1.5em}
\end{figure}

To bridge the substantial gap between self-supervised and supervised representation learning on fine-grained object recognition, we first analyze and compare knowledge learned by various self-supervised methods and supervised methods during pre-training. We find that current self-supervised contrastive learning methods tend to learn low-level texture information and lack the localization ability of the foreground object. In contrast, the supervised method shows better localization ability. Specifically, we show that the incompetence of localization of current contrastive learning is primarily due to the commonly adopted \textit{RandomResizedCrop} (RRC) augmentation, where a random size patch at a random location is cropped and resized to the original size. The model then might learn a semantic representation of the bird by contrasting the tree and the wing of the bird, as illustrated in Figure~\ref{fig_1}. This practice may be reasonable for coarse-grained recognition if background cues are more associated with the class than the foreground cues (e.g. $p(bird|tree)> p(car|tree)$). However, the background of the image being a tree is not as informative when distinguishing bird species. Consequently, the model learns by cheating on picking low-level texture clues (usually from the background) instead of learning by localizing the foreground. This phenomenon is mutual for existing contrastive methods such as MoCo.v2\cite{2020mocov2}, BYOL\cite{nips2020byol} despite different contrastive mechanisms.

The devil lies in semantically discriminative fine-grained feature extraction for a successful contrastive pre-training. To remedy the inadequacy of fine-grained feature capturing due to failure in localizing to discriminative regions, we propose to empower contrastive learning with localization ability by aligning fine-grained semantic features across augmented views. In particular, we come up with a pre-training framework called Cross-View Saliency Alignment (CVSA). CVSA consists of two algorithmic components: (a) A general plug-and-play data augmentation strategy called SaliencySwap, which swaps the saliency region of the reference image with the saliency region of a randomly selected background image. SaliencySwap ensures semantic consistency between augmented views while introducing background variation. A demonstration of SaliencySwap in comparison with RRC is shown in Figure~\ref{fig_1}. An alignment loss that provides an explicit localization supervision signal by forcing the model to give the highest correspondence response intensity of the foreground object across views.

On top of the proposed CVSA, to further bridge the performance gap between self-supervised and supervised representation learning on the fine-grained recognition problem, we offer a \textit{dual-stage} pre-training setting, which utilizes coarse-grained datasets for low-level feature extraction and fine-grained datasets for high-level target discrimination and localization.
In short, this paper makes the following contributions:


\begin{itemize}
\vspace{-0.5em}
\item We delve deep into the knowledge learned by various self-supervised methods compared to supervised methods during the fine-grained pre-training phase and point out the cause of limitations.
\vspace{-0.75em}
\item We develop a novel contrastive learning framework for fine-grained recognition, which contains a data augmentation technique called \emph{SaliencySwap} to guarantee semantic consistency between views and an \emph{alignment objective} which enables the model to localize.
\vspace{-0.75em}
\item Extensive experiments show consistent performance gain of CVSA under various pre-training stage settings on small- and large-scale fine-grained benchmarks.
\end{itemize}

\section{Cross-view Saliency Alignment}
\label{sec: cvsa}
We explore the capabilities learned out of three classes of pre-training mechanisms, namely self-supervised contrastive, non-contrastive, and supervised methods. In particular, we focus on discriminative feature extraction and object localization ability. Without loss of generality, we select MoCo.v2, BYOL, Rot-Pred, and supervised classification for comparison. We discover that compared to supervised methods, self-supervised methods show \textbf{worse object localization ability} and \textbf{discriminative feature extraction ability} is also crucial for fine-grained categorization. The details of the experiments and corresponding analysis are provided in Appendix~\ref{sec_empirical_study}. Based on our observations, given a fine-grained classification problem, similar to \cite{arora2019theoretical}, we assume $\mathcal{X}$ to be a set of all samples with an underlying set of discrete latent classes $\mathcal{C}$ that represent semantic content, we obtain the joint distribution between each sample $x$ and its class $c$:
\vspace{-0.5em}
\begin{align}
    \vspace{-0.3em}
    p(c,x) = p(c|x_{fore}) \cdot p(x_{fore}|x),
    \vspace{-0.3em}
\end{align}
where $x_{fore}$ stands for the foreground object. The intuition behind this factorization suggests that given an image of a fine-grained object, the model should localize the foreground object ($p(x_{fore}|x)$) to discriminate the species of the foreground object ($p(c|x_{fore})$). Following this formulation, we propose a \textit{dual-stage} pre-training pipeline for self-supervised fine-grained recognition with the \textit{first-stage} learning discriminative texture extraction ability and \textit{second-stage} learning localization capability.

\begin{figure*}[ht]
  \centering
  \includegraphics[width=0.95\linewidth]{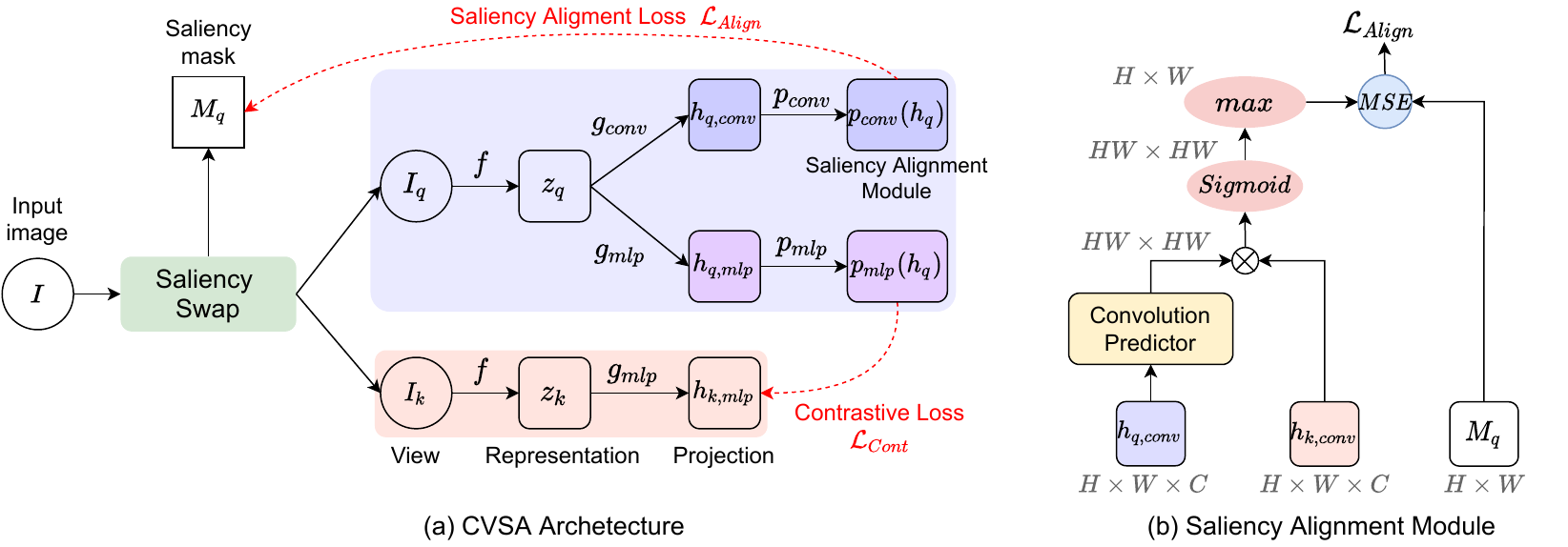}
  \vspace{-1em}
  \caption{Learning paradigm of Cross-view Saliency Alignment. (a) The network parameters in red are an exponential moving average of the purple part, and the contrastive loss $\mathcal{L}_{Cont}$ is calculated between $p_{q,mlp}$ and $h_{k,mlp}$ (stop-gradient) as BYOL. (b) Saliency alignment module calculates $\mathcal{L}_{Align}$ between predicted attention maps and input saliency masks $M$.}
  \label{sytem_fig}
\vspace{-1.0em}
\end{figure*}

\subsection{SaliencySwap}
\label{subsec:ss}
SaliencySwap maximally utilizes the saliency information for foreground semantic consistency across views while introducing background variation. SaliencySwap guarantees that each view at least contains part of the foreground object and thus prevents the encoder from learning irrelevant feature representation through pure background information.

\subsubsection{Source Saliency Detection}
A saliency detection algorithm generates a saliency map that indicates the objects of interest (primarily foreground). Let $I \in \mathbb{R}^{W \times H \times C}$ be an image in the training set, define $\psi$ to be a saliency detection algorithm, then the output saliency map $S_{i,j}=\psi(I_{i,j}) \in \mathbb{R}^{W \times H}$ indicates the saliency intensity value at pixel $I_{i,j}$. The saliency information can be noisy. Therefore, we seek to find a bounding box $B=(l, t, W_b, H_b)$ of the foreground object with the highest averaged saliency information satisfying the following objective function:
\vspace{-0.5em}
\begin{align}
    \underset{W_b, H_b, l, t}{\mathrm{argmax}}
    \sum_{i=l}^{i=l+W_b} \sum_{j=t}^{j=t+H_b}
    \frac{S_{i,j}}{W_b \times H_b}.
    \label{eqation:2}
\end{align}
A corresponding binary saliency mask $M\in \mathbb{R}^{W\times H}$ is defined by filling with 1 within the bounding box $B$, otherwise 0. Then we crop a random patch within the bounding box $B$. Similar to RRC, the size of the patch is determined based on an area ratio (to the area of the bounding box), which is sampled from a uniform distribution $U(\lambda,1)$.

\subsubsection{Foreground Background Fusion}
We then combine the cropped foreground patch from the source image (foreground image) with another randomly selected image (background image). To avoid saliency ambiguity, we restrict each augmented view to contain the saliency information only of one semantic object. We consider two ways of merging: (I) The background dataset is the same as the foreground dataset. The saliency information of the background needs to be eliminated. We first calculate the bounding box $B_f$, $B_b$ of the foreground and background images, respectively, using Eqn. \ref{eqation:2}. For scenic images, we use the bounding box with maximum area. Then select a random patch from the foreground $B_f$ and resize it to the shape of $B_b$. Finally, we replace $B_b$ with the resized foreground patch. (II) The background dataset is different from the foreground dataset. We choose a dataset like COCO rather than the iconic dataset like IN such that the background dataset is rich in environments. Again, we calculate the bounding box $B_f$ of the foreground and select a random patch. Then we resize the selected patch based on an area ratio (to the area of the background) which is sampled from a uniform distribution $(\beta,1)$. Finally, we 'paste' the resized patch to a random location in the background.

\subsection{Cross-view Saliency Alignment}
\label{sec:CAM}
Given two views $I_q$ and $I_k$ of Image $I$ augmented by a pipeline containing SaliencySwap and other augmentation operations such as random flipping and color jittering. Define $M_k$ and $M_q$ to be their saliency masks, respectively. Let $z^{l}_q$ and $z^{l}_k \in \mathbb{R}^{W^l\times H^l\times C^l}$ be the $C^l$ dimensional $H^l\times W^l$ feature maps encoded by an encoder network $z=f(I)$ (\textit{e.g.,} ResNet) truncated at stage $l$. We adopt two types of non-linear projector necks $g(\cdot)$ on top of the encoder to form a $d$-dim projection. The two-layer MLP projector generates the following projection $h_{mlp}=g_{mlp}(z)$ as proposed by SimCLR. Similarly, a convolutional projection $h_{conv}$ is given by a convolution projector $g_{conv}$, which consists of two $1\times 1$ convolution layers with a batch normalization layer and a ReLU layer in between. Following BYOL, two predictor heads $p_{mlp}(\cdot)$ and $p_{conv}(\cdot)$, which have the same network structures except for different input dimensions, are adopted to match the output of one view to the other. Specially, $p_{conv}(\cdot)$ is designed for saliency alignment. Figure~\ref{sytem_fig} (a) shows the overall framework.

\textbf{Cross-view Attention.}\quad
We seek to capitalize on the pixel-level foreground semantic interactions between the feature maps of two different augmented views. We first build a cross-view attention map:
\vspace{-1.25em}
\begin{align}
    \vspace{-0.2em}
    A^l_{q,k} = p_{conv}(h^{l}_{q,conv}) \otimes {h^{l}_{k,conv}}^{T},
    \vspace{-0.2em}
\label{eq:3}
\end{align}
where $A^l_{q,k}$ denotes the attention map is of view $k$ w.r.t. view $q$, $T$ denotes matrix transposition, and $\otimes$ denotes matrix multiplication. The location-aware attention map $A^l\in \mathbb{R}^{W^{l}H^{l}\times H^{l}W^{l}}$ indicates a pair-wise spatial correspondence between any pixel from $p_{conv}(h^{l}_{q,conv})$ and any pixel from $h^{l}_{k,conv}$. Symetrically, we get the attention map $A^l_{k,q}$ of view $q$ w.r.t. view $k$ by interchanging $q$ and $k$ of Eqn.~\ref{eq:3}.

\textbf{Joint Saliency Alignment.}\quad
To enhance the encoder's ability to identify the location of the foreground object, we propose to align the saliency mask with a correspondence intensity matrix that captures the pixel-level correlation from the feature map of one view to the other, as shown in Figure~\ref{sytem_fig} (b). The correspondence intensity matrix $C$ is formulated as follows:
\vspace{-0.5em}
\begin{align}
    \vspace{-0.2em}
    C_{q,k}^{l} = {\rm{max}}(\sigma(A_{q,k}^{l})),
    \vspace{-0.2em}
\end{align}
where $\sigma(\cdot)$ denotes sigmoid activation. Note that the shape of $C_{q,k}^{l}$ is $W^l\times H^l$ and the $max(\cdot)$ operation is performed over the second axis of $A_{q,k}^{l}$.
We then define a symmetrized alignment loss between the saliency mask $M$ and the correspondence intensity matrix $C$:
\vspace{-0.5em}
\begin{align}
    \vspace{-0.2em}
    \mathcal{L}_{Align} = \left\|\delta^{l}(M_{q}) - C^{l}_{q,k}\right\|^2 + \left\| \delta^{l}(M_{k})-C^{l}_{k,q}\right\|^2,
    \vspace{-0.2em}
\end{align}
where $\delta^{l}(\cdot): \mathbb{R}^{W\times H} \rightarrow \mathbb{R}^{W^l\times H^l}$ denotes the bilinear downsampling operation. The proposed alignment loss restricts the most cross-view correlated pixels to the saliency region and thus gives the model localization ability. In addition, leveraging cross-layer semantics also enhances the representation of multi-scale learning.

\textbf{Joint Objective.}\quad
We define a contrastive loss ${\mathcal{L}}_{Cont}$ with the prediction vector $p_{q,mlp} \stackrel{\text{def}}{=} p_{mlp}(h_{q,mlp})$ and the projection vector $h_{k,mlp}$ using negative cosine similarity $\mathcal{D}(\cdot)$ as:
\vspace{-0.5em}
\begin{align}
    \vspace{-0.2em}
    \mathcal{L}_{Cont} = \frac{1}{2}\mathcal{D}(p_{q,mlp}, h_{k,mlp}) + \frac{1}{2}\mathcal{D}(p_{k,mlp}, h_{q,mlp}),
    \vspace{-0.2em}
\end{align}
where $\mathcal{D}(p, h) = -\frac{p}{\left\|{p}\right\|_2} \cdot \frac{h}{\left\|{h}\right\|_2}$. The joint objective for the \textit{second-stage} pretext task is:
\vspace{-0.6em}
\begin{align}
    \vspace{-0.20em}
    \mathcal{L}_{CVSA} = \mathcal{L}_{Cont} + \mathcal{L}_{Align}.
    \vspace{-0.2em}
\end{align}

\vspace{-0.5em}
\section{Experiments}
\label{sec_exp}

\subsection{Experimental settings}
\textbf{Implementation details.}\quad
We use ResNet as the encoder $f$ followed by two projectors and two predictor sub-networks. During the \textit{first-stage} pre-training, we follow the exact experimental setup in BYOL. For the \textit{second-stage} in \textit{dual-stage} pre-training, the model is initialized with the pre-trained weight from the \textit{first-stage} and the first two stages of the ResNet backbone are frozen. We use a learning rate of $lr\times BatchSize/256$ with $BatchSize=1024$ and a base $lr$ selected from $\{0.3, 0.6, 0.9, 1.2\}$. The embedding dimension is set to $d=256$ as BYOL. As for image augmentations, we follow the settings in MoCo.v2 for all contrastive learning methods. We replace \textit{RandomResizedCrop} (RRC) with the proposed SaliencySwap (SS) and adopt all other augmentations in MoCo.v2~\cite{2020mocov2}, detailed in Appendix~\ref{sec_empirical_study}. We grid search cropping scale ratio of SaliencySwap $\lambda \in \{0.08, 0.2, 0.5, 0.8\}$ and set $\lambda=0.5$ by default. To balance the performance and computational cost, we adopt the stage $l=4$ for the alignment. For simplicity, we use the ground truth bounding box of fine-grained datasets because most saliency detectors are trained in a supervised manner. 
An ablation study of the performance using different detection algorithms is given in Appendix~\ref{app_ablation_detection}.
We use the same pre-training setup as in Appendix~\ref{sec_settings} for other unstated setups.

\textbf{Dataset.}\quad
We assess the performance of the representation pre-trained using \textit{dual-stage}, \textit{first-stage} only and \textit{second-stage only} on four small-scale and one large-scale fine-grained benchmarks: 1) CUB-200-2011~\cite{wah2011caltech} (CUB) contains 11,788 images from 200 wild bird species, 2) Stanford-Cars~\cite{krause20133d} (Cars) contains 16,185 images of 196 car subcategories, 3) FGVC-Aircraft~\cite{maji2013fine} (Aifcrafts) contains 10,000 images of 100 classes of aircrafts, 4) NA-birds~\cite{van2015building} (NAbirds) is a large dataset with 48,562 images for over 555 bird classes and 5) iNaturalist2018 (iNat2018)~\cite{cvpr2018inaturalist} is an unbalanced long tail dataset which contains 437,513 images of 8,142 taxa coming from 14 super-classes. We follow the standard dataset partition in the original works. For the \textit{first-stage} pre-training, we adopt two popular datasets: 1) ImageNet-1k (IN-1k) contains 1.28 million of training images. 2) MS-COCO (COCO) contains 118k images with more complex scenes of many objects.


\textbf{Hyperparameters for CVSA.}\quad
We search for the optimal hyperparameters for our proposed CVSA on the validation set of fine-grained datasets using ResNet-18 as the encoder. We set the batch size to $1024$ and follow other training settings of BYOL~\cite{nips2020byol}. The base learning rate is set to $0.3$ for the Cars dataset and $0.6$ for the other three datasets. The cropping scale ratio of SaliencySwap is $\lambda=0.5$ by default. As for the stage $l$ in CVSA, we analyze the performance and the computational cost of using $l\in \{3,4\}$ since the first two stages are frozen in the \textit{dual-stage}. Furthermore, we find that using $l=4$ achieves a balance between the performance and the computation cost.

\subsection{Comparison with State-of-the-art}
We choose two hand-crafted methods (Rel-Loc and Rot-Pred), three commonly used contrastive learning methods (SimCLR, MoCo.v2, and BYOL), and three extension methods (LooC*, DiLo, and InsLoc) for comparison. DiLo and InsLoc are reproduced using official code, with other methods reproduced using OpenMixup~\cite{li2022openmixup}. LooC* denotes its rotation version reproduced by us. Since our approach extends BYOL, we choose BYOL as the baseline. Notice that \gray{\bf{grey}} indicates baseline, \green{\bf{green}} denotes improvement over baseline while \red{\bf{red}} for degradation performance, \textbf{bold} denotes the best performance.

\begin{figure*}[t]
\vspace{-1em}
\begin{minipage}{0.49\linewidth}
\centering
    \begin{table}[H]
    \centering
    \setlength{\tabcolsep}{0.9mm}
    \resizebox{1\columnwidth}{!}{
    \begin{tabular}{lcccc}
    \toprule
    Methods            & CUB                   & NAbirds               & Aircrafts             & Cars  \\
    \hline
    \gray{Random}      & 58.51 \redbf{-6.34}   & 65.78 \redbf{-6.76}   & 70.58 \redbf{-2.02}   & 70.51 \redbf{-5.36} \\
    Rel-Loc            & 65.89 \gbf{+1.04}     & 72.60 \gbf{+0.06}     & 72.24 \redbf{-0.36}   & 75.27 \redbf{-0.60} \\
    Rot-Pred           & 66.67 \gbf{+1.82}     & 73.01 \gbf{+0.47}     & 72.67 \gbf{+0.07}     & 75.48 \redbf{-0.39} \\
    SimCLR             & 63.43 \redbf{-1.42}   & 72.05 \redbf{-0.49}   & 72.42 \redbf{-0.18}   & 75.12 \redbf{-0.75} \\
    MoCo.v2            & 63.21 \redbf{-1.64}   & 71.36 \redbf{-1.18}   & 71.93 \redbf{-0.67}   & 74.89 \redbf{-0.98} \\
    LooC*              & 66.42 \gbf{+1.57}     & 72.84 \gbf{+0.30}     & 72.49 \redbf{-0.11}   & 75.69 \redbf{-0.18} \\
    InsLoc             & 64.87 \gbf{+0.02}     & 72.80 \gbf{+0.26}     & 73.43 \gbf{+0.83}     & 76.61 \gbf{+0.64} \\
    BYOL               & 64.85 \rbf{+0.00}     & 72.54 \rbf{+0.00}     & 72.60 \rbf{+0.00}     & 75.87 \rbf{+0.00} \\
    BYOL+DiLo          & 66.16 \gbf{+1.31}     & 73.12 \gbf{+0.58}     & 73.52 \gbf{+0.92}     & 76.36 \gbf{+0.49} \\
    \textbf{BYOL+CVSA} & \bf{66.88} \gbf{+2.03} & \bf{73.75} \gbf{+1.21} & \bf{74.55} \gbf{+1.95} & \bf{77.45} \gbf{+1.58} \\
    \bottomrule
    \end{tabular}
    }
    \vspace{-0.75em}
    \caption{\textbf{Comparison of \textit{second-stage only} pre-training on fine-grained benchmarks.} Top-1 accuracy (\%) under fine-tuned evaluation is reported. Random denotes random initialized models in the \textit{second-stage}. BYOL is regarded as the baseline by default.}
    \label{tab:stage1}
    \vspace{-1em}
\end{table}

\end{minipage}
\begin{minipage}{0.51\linewidth}
\centering
    \begin{table}[H]
    \centering
    \setlength{\tabcolsep}{0.9mm}
    \resizebox{1\columnwidth}{!}{
    \begin{tabular}{lccccc}
    \toprule
    Methods            & Stage 2    & CUB                   & NAbirds               & Aircrafts             & Cars  \\
    \hline
    Rel-Loc            & \checkmark & 67.33 \redbf{-1.22}   & 73.82 \redbf{-2.47}   & 81.20 \redbf{-0.53}   & 85.80 \redbf{-0.56} \\
    Rot-Pred           & \checkmark & 67.75 \redbf{-0.80}   & 74.26 \redbf{-2.03}   & 81.58 \redbf{-0.15}   & 85.74 \redbf{-0.62} \\
    SimCLR             & $\times$   & 68.30 \redbf{-0.30}   & 73.51 \redbf{-2.78}   & 81.18 \redbf{-0.55}   & 85.93 \redbf{-0.43} \\
    MoCo.v2            & $\times$   & 68.47 \redbf{-0.08}   & 73.73 \redbf{-2.56}   & 81.78 \gbf{+0.05}     & 86.08 \redbf{-0.28} \\
    MoCo.v2            & \checkmark & 67.60 \redbf{-1.05}   & 73.18 \redbf{-3.11}   & 81.04 \redbf{-0.69}   & 85.71 \redbf{-0.65} \\
    LooC*              & \checkmark & 68.71 \gbf{+0.16}     & 74.45 \redbf{-1.84}   & 81.75 \gbf{+0.02}     & 85.90 \redbf{-0.46} \\
    InsLoc             & \checkmark & 67.94 \redbf{-0.20}   & 76.36 \gbf{+0.07}     & 81.54 \redbf{-0.23}   & 86.38 \gbf{+0.02} \\
    BYOL               & $\times$   & 68.55 \rbf{+0.00}     & 76.29 \rbf{+0.00}     & 81.73 \rbf{+0.00}     & 86.36 \rbf{+0.00} \\
    BYOL               & \checkmark & 68.01 \redbf{-0.54}   & 75.82 \redbf{-0.47}   & 80.53 \redbf{-1.20}   & 85.49 \redbf{-0.87} \\
    BYOL+DiLo          & \checkmark & 68.70 \gbf{+0.15}     & 76.94 \gbf{+0.65}     & 82.04 \gbf{+0.31}     & 86.46 \gbf{+0.10} \\
    \textbf{BYOL+CVSA} & \checkmark & \bf{69.14} \gbf{+0.59} & \bf{77.57} \gbf{+1.28} & \bf{82.77} \gbf{+1.04} & \bf{87.13} \gbf{+0.77} \\
    \bottomrule
    \end{tabular}
    }
    \vspace{-0.75em}
    \caption{\textbf{Comparison of \textit{dual-stage} pre-training on fine-grained benchmarks.} Top-1 accuracy (\%) under fine-tuned evaluation is reported. The \textit{first-stage} is pre-trained on COCO, and \checkmark denotes performing the \textit{second-stage} pre-training on fine-grained datasets.}
    \label{tab:stage2_coco}
    \vspace{-1em}
\end{table}

\end{minipage}
\vspace{-1.25em}
\end{figure*}

\begin{wraptable}{r}{0.55\textwidth}
    \setlength{\tabcolsep}{0.9mm}
    \resizebox{0.55\columnwidth}{!}{
    \begin{tabular}{lccccc}
    \toprule
    Methods            & Stage 2    & CUB                   & NAbirds               & Aircrafts             & Cars  \\
    \hline
    \gray{Supervised}  & $\times$   & \gray{81.02}          & \gray{80.09}          & \gray{87.25}          & \gray{90.61} \\
    SimCLR             & $\times$   & 73.99 \redbf{-2.64}   & 76.30 \redbf{-2.59}   & 85.96 \redbf{-1.23}   & 88.16 \redbf{-1.43} \\
    MoCo.v2            & $\times$   & 73.19 \redbf{-3.44}   & 75.64 \redbf{-3.25}   & 85.49 \redbf{-1.70}   & 87.51 \redbf{-2.08} \\
    MoCo.v2            & \checkmark & 71.77 \redbf{-4.86}   & 73.96 \redbf{-4.93}   & 83.25 \redbf{-3.94}   & 86.88 \redbf{-2.71} \\
    InsLoc             & \checkmark & 75.83 \redbf{-0.80}   & 78.86 \redbf{-0.03}   & 86.60 \redbf{-0.59}   & 88.87 \redbf{-0.72} \\
    BYOL               & $\times$   & 76.63 \rbf{+0.00}     & 78.89 \rbf{+0.00}     & 87.19 \rbf{+0.00}     & 89.59 \rbf{+0.00} \\
    BYOL               & \checkmark & 72.46 \redbf{-4.17}   & 76.12 \redbf{-2.77}   & 84.58 \redbf{-3.61}   & 87.08 \redbf{-2.51} \\
    BYOL+DiLo          & \checkmark & 76.60 \redbf{-0.03}   & 79.04 \gbf{+0.15}     & 87.03 \redbf{-0.16}   & 89.26 \redbf{-0.33} \\
    \textbf{BYOL+CVSA} & \checkmark & \bf{77.10} \gbf{+0.47} & \bf{79.64} \gbf{+0.75} & \bf{87.27} \gbf{+0.12} & \bf{89.76} \gbf{+0.18} \\
    \bottomrule
    \end{tabular}
    }
    \vspace{-0.75em}
    \caption{\textbf{Comparison of \textit{dual-stage} pre-training on fine-grained benchmarks.} Top-1 accuracy (\%) under fine-tuned evaluation is reported. The \textit{first-stage} is pre-trained on IN-1k, and \checkmark denotes performing the \textit{second-stage} pre-training on corresponding fine-grained datasets. Supervised denotes the supervised pre-training on IN-1k.}
    \label{tab:stage2_IN}

    \vspace{-1.10em}
\end{wraptable}

\textbf{Small-scale scenarios.}\quad
We perform 800 epochs pre-training with ResNet-50 encoder using different pre-training stage settings on four small-scale fine-grained datasets and report the top-1 classification accuracy under the fine-tune protocol. 
As shown in Table~\ref{tab:stage1}, when performing the \textit{second-stage only} setting, namely pre-training with the training set of fine-grained datasets from scratch, our proposed CVSA outperforms BYOL by a large margin on all benchmarks. When applying the \textit{dual-stage} setting using COCO for the \textit{first-stage}, as shown in Table~\ref{tab:stage2_coco}, CVSA shows a consistent improvement on representations learned during the \textit{first-stage} while most comparing methods yield worse performance than the \textit{first-stage} pre-trained BYOL baseline. However, when using ImageNet for the \textit{first-stage} pre-training, as shown in Table~\ref{tab:stage2_IN}, the improvement of CVSA in the \textit{dual-stage} setting is not as significant on Aircraft and Cars. We hypothesize that this is due to the iconic nature of these two datasets. We find that most images of these two datasets are of similar and straightforward backgrounds, restricting the background variation imposed by our approach. Besides, comparison of Table~\ref{tab:stage1} with Table~\ref{tab:stage2_coco} and~\ref{tab:stage2_IN} suggests that the performance gain is affected by the size of dataset used. This is consistent with our formulation in Appendix~\ref{sec_formulation} that discriminative feature extraction relies on the size of the pre-training dataset, and localization ability alone is not enough for fine-grained recognition. In a word, CVSA and \textit{dual-stage} pre-training pipeline improve learned representation for the small-scale fine-grained classification.

\begin{wraptable}{r}{0.60\textwidth}
    \vspace{-1.1em}
    \resizebox{0.60\textwidth}{!}{
    \setlength\tabcolsep{3pt} 
    \begin{tabular}{ccccccc}
        \toprule
        Stage 1  & Stage 2  & Sup.        & MoCo.v2           & BYOL          & BYOL+DiLo & \bf{BYOL+CVSA}            \\ \hline
        $\times$ & iNat2018 & \gray{5.7}  & 41.8\redbf{-2.8} & 44.6\rbf{+0.0} & 44.3\redbf{-0.3} & \bf{45.4}\gbf{+0.8} \\
        IN-1k    & iNat2018 & \gray{46.5} & 45.0\redbf{-1.9} & 46.9\rbf{+0.0} & 47.0\gbf{+0.1}   & \bf{47.5}\gbf{+0.6} \\
        \bottomrule
    \end{tabular}}
    \vspace{-0.75em}
    \caption{\textbf{Comparison of \textit{dual-stage} pre-training on iNat2018.} Top-1 accuracy (\%) under linear evaluation is reported. \textit{Sup.} denotes the supervised pre-training on iNat2018 in the \textit{second-stage}.}
    \label{tab:inat}

    \vspace{-1.5em}
\end{wraptable}

\textbf{Large-scale scenarios.}\quad
Next, we evaluate CVSA on large-scale benchmark iNat2018 with the \textit{dual-stage} pre-training. Following MoCo~\cite{cvpr2020moco}, we adopt the linear evaluation protocol as mentioned in Appendix~\ref{sec_settings} training 100 epochs with the basic learning rate $lr=0.025$ and batch size of $256$. In Table~\ref{tab:inat}, CVSA outperforms existing methods in both settings indicating its effectiveness in the large-scale datasets, especially improving the baseline by 0.9\% with \textit{second-stage-only}. Compared with the results in Table~\ref{tab:stage2_IN}, the performance gain is more significant than \textit{second-stage} pre-training on small-scale datasets like Aircraft and Cars, which suggests that CVSA benefits from a larger data size. Meanwhile, using \textit{dual-stage} pre-training with IN-1k for the \textit{first-stage} helps contrastive methods to learn better representations than the \textit{second-stage only} setting.

\subsection{Ablation Study}
\label{subsec:ablation}
We perform 400 epochs pre-training with ResNet-18 on CUB for the first three ablation studies. As for the fourth ablation for the \textit{dual-stage} pre-training, we perform 200 epoch \textit{first-stage} pre-training on large datasets and 400 epochs \textit{second-stage} pre-training on target datasets. The top-1 accuracy under the fine-tune evaluation is reported. Appendix~\ref{app_ablation_detection} provides more results.

\begin{figure*}[t]
\begin{minipage}{0.55\linewidth}
\centering
    \begin{figure}[H]
    \vspace{-1em}
        \centering
        \includegraphics[width=\linewidth]{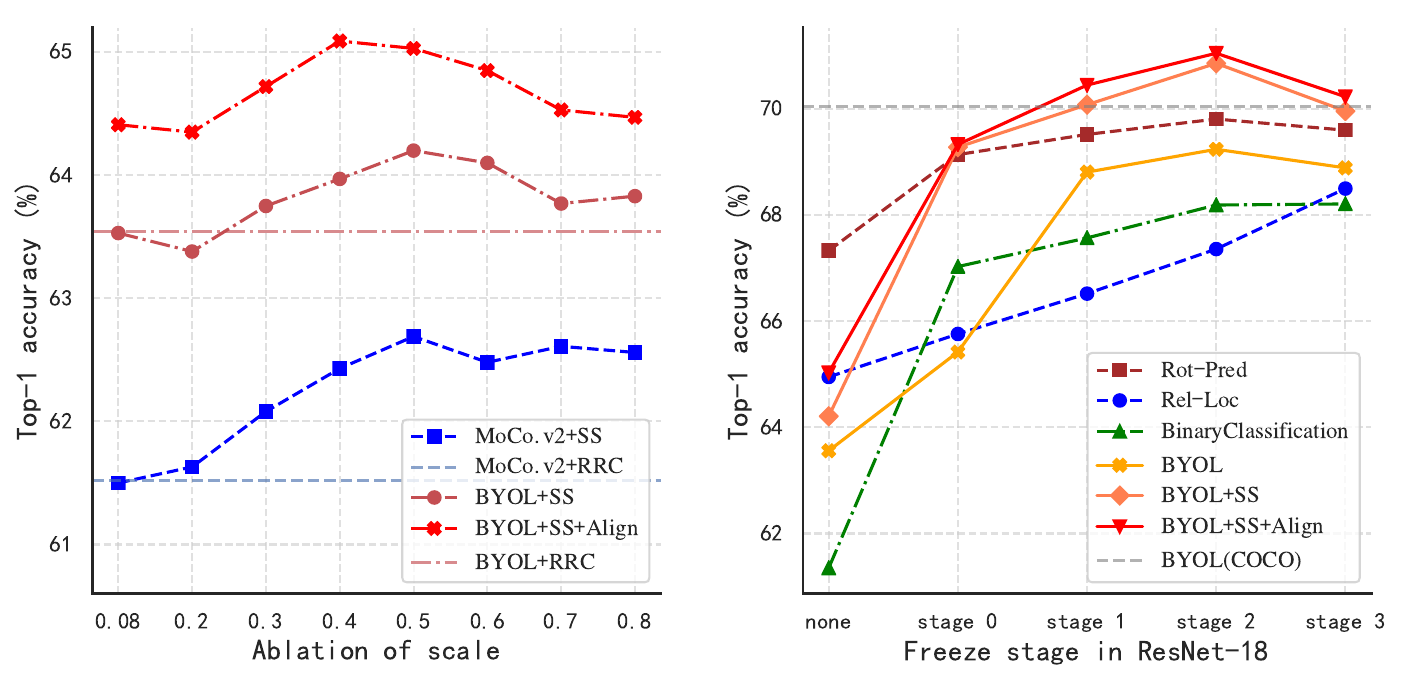}
        \vspace{-2em}
        \caption{\textbf{Ablation of hyperparameters}. Left: the effect of different scaling factors of SS during \textit{second-stage}. Right: the effect of freezing different ResNet stages during \textit{second-stage}. The dotted grey line indicates the performance of BYOL pre-trained on COCO during \textit{first-stage}.
        }
        \label{fig:ss_scale_ablation}
    \end{figure}
\end{minipage}
\begin{minipage}{0.44\linewidth}
\centering
    \begin{table}[H]
    \centering
\resizebox{1.0\linewidth}{!}{
\setlength\tabcolsep{3pt} 
    \begin{tabular}{lccccc}
    \toprule
    Methods        & Stage 1  & Stage 2      & CUB       & Aircrafts & Cars      \\ \hline
    BYOL           & IN-1k    & $\times$     & \bf{76.6} & 87.2      & 89.6      \\
    \bf{BYOL+CVSA} & IN-1k    & $\times$     & 75.9      & 86.8      & 89.2      \\
    \bf{BYOL+CVSA} & IN-1k    & $\checkmark$ & 76.5      & \bf{87.3} & \bf{89.7} \\ \hline
    BYOL           & iNat2018 & $\times$     & \bf{77.2} & 80.4      & 85.8      \\
    \bf{BYOL+CVSA} & iNat2018 & $\checkmark$ & 77.0      & \bf{81.2} & \bf{86.5} \\ \hline
    BYOL           & COCO     & $\times$     & 68.6      & 81.7      & 86.4      \\
    \bf{BYOL+CVSA} & COCO     & $\checkmark$ & \bf{68.9} & \bf{82.3} & \bf{86.7} \\
    \bottomrule
    \end{tabular}
    }
    \vspace{-0.75em}
    \caption{\textbf{Ablation of \textit{dual-stage} pre-training.} The \textit{first-stage} is pre-trained 200 epochs on IN-1k and iNat2018 and 400 epochs on COCO. The \textit{second-stage} is pre-trained on downstream datasets (denoted by $\checkmark$) or not (denoted by $\times$).
    }
    \label{tab:ablation_dual}
    \vspace{-2em}
\end{table}

\end{minipage}
\vspace{-1.5em}
\end{figure*}

\textbf{Module effectiveness ablation.}\quad
We demonstrate the effectiveness of our CVSA by adding modules one by one onto the baseline. We compare the performance of MoCo.v2 and BYOL using \textbf{SS} against \textit{RandomResizedCrop} (RRC) while keeping all other augmentations unchanged. As shown in Figure~\ref{fig:ss_scale_ablation} left, \textbf{SS} outperforms RRC (using the default scaling factor of $0.08$) both for MoCo.v2 and BYOL. We observe a further performance improvement from adding the saliency alignment loss (\textbf{Align}) onto BYOL using \textbf{SS}. Now, we have shown that both \textbf{SS} and \textbf{Align} contribute to higher performance.

\textbf{Hyperparameter ablation.}\quad
We then analyze the performance of \textbf{SS} using different crop scaling factors, specifically $\lambda \in \{0.08, 0.2, 0.3, 0.4, 0.5, 0.8\}$ in Figure~\ref{fig:ss_scale_ablation} left. From Figure~\ref{fig:analysis} left, the performance of BYOL fluctuates drastically under the different scaling factors of RRC, with the best result achieved with a scaling factor being $0.08$. However, BYOL yields similar performance under different choices of $\lambda$ of \textbf{SS}. We argue that \textbf{SS}, together with \textbf{Align}, helps the model to localize on the foreground object, and thus local texture is no longer the only clue for contrastive pre-training to learn representation. Then, we study the effect when freezing various stages of ResNet-18 in the \textit{second-stage}. We initialize the model of all methods in the \textit{second-stage} with the weights of BYOL baseline pre-trained on COCO in the \textit{first-stage}. In Figure~\ref{fig:ss_scale_ablation} right, the horizontal axis indicates freezing up to different stages of ResNet-18. The best performance is reached when freezing up to the \textit{second-stage} of ResNet. The early stages of ResNet mostly extract low-level texture information, while the later stages are for higher-level features such as discrimination and localization. Intuitively, we wish to enhance the localization ability without jeopardizing the texture extraction ability acquired during the \textit{first-stage}. This explains the reason for freezing the first two stages during the \textit{second-stage}, which is common in object detection~\cite{NIPS2015_14bfa6bb}.

\begin{wraptable}{r}{0.50\textwidth}
    \vspace{-1.0em}
    \centering
    \setlength\tabcolsep{3pt} 
    \resizebox{0.50\textwidth}{!}{
        \begin{tabular}{lccc}
        \toprule
        Background  & BYOL+DiLo             & BYOL+SS        & BYOL+SS+Align \\
        \hline
        CUB         & 64.14 \rbf{+0.00}     & \bf{64.35} \rbf{+0.00} & 65.02 \rbf{+0.00}  \\
        COCO        & 60.07 \redbf{-4.07}   & 60.42 \redbf{-3.93}   & 61.23 \redbf{-3.79}  \\
        NAbirds     & 62.51 \redbf{-1.81}   & 63.06 \redbf{-1.29}   & 64.80 \redbf{-0.22}  \\
        CUB+COCO    & 61.26 \redbf{-2.88}   & 61.49 \redbf{-2.86}   & 62.01 \redbf{-3.01} \\
        CUB+NAbirds & \bf{64.28}\gbf{+0.14} & 64.23 \redbf{-0.12}   & \bf{65.50}\gbf{+0.48} \\
        \bottomrule
    \end{tabular}}
    \vspace{-0.75em}
    \label{tab: table4}
    \caption{\textbf{Evaluation of background datasets extension for the \textit{second-stage} pre-training.} We study the effect of fusing different background datasets on CUB.
    }

    \vspace{-1.0em}
\end{wraptable}

\textbf{Background dataset ablation.}\quad
Moreover, we compare different foreground and background fusion methods (\textbf{SS} and DiLo) based on CUB dataset using \textit{second-stage only} settings. As shown in Table~\ref{tab: table4}, replacing the background with other datasets usually results in performance degradation, while fusing with complex backgrounds might improve the performance. We hypothesize that exploring task-specific backgrounds is more critical in fine-grained scenarios. Meanwhile, it indicates that the proposed CVSA (\textbf{SS}+\textbf{Align}) can well explore the background of CUB.

\textbf{First-stage pre-training ablation.}\quad
Additionally, we verify the necessity of the \textit{first-stage} learning discrimination on iconic datasets in the \textit{dual-stage} pre-training scheme. We compare naive BYOL and CVSA for the \textit{first-stage} pre-training on IN-1k, iNat2018, and COCO. As shown in Table~\ref{tab:ablation_dual}, using IN-1k (iconic) for the \textit{first-stage} yields better performance that scenic datasets, indicating the necessity of the \textit{first-stage} on iconic datasets. Since CVSA is design for the \textit{second-stage} on target datasets, we find that using CVSA on \textit{second-stage} outperforms naive BYOL while producing degraded performance on the \textit{first-stage}.

\begin{wrapfigure}{r}{0.60\textwidth}
    \centering
    \vspace{-1.0em}
    \includegraphics[width=1.0\linewidth]{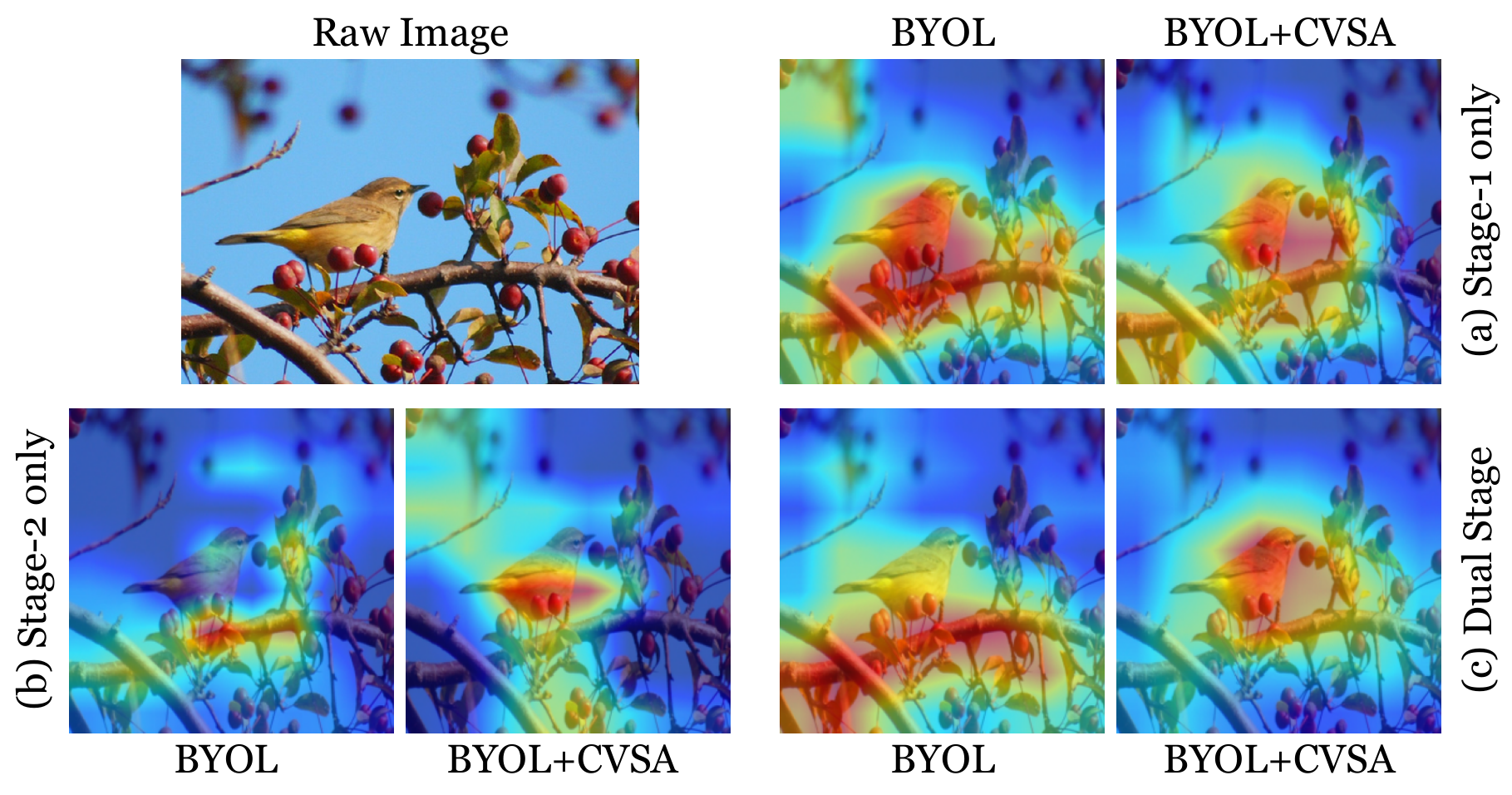}
    \vspace{-2.0em}
    \caption{Grad-CAM visualization of BYOL and BYOL+CVSA on CUB200.
    }
    \label{fig:vis_stage}
    \vspace{-1.0em}
\end{wrapfigure}

\textbf{Visualization of localization abilities.}\quad
Lastly, we compare the localization abilities by Grad-CAM~\cite{iccv2017gradcam} visualization of SS (BYOL+CVSA) and RRC (BYOL). Figure~\ref{fig:vis_stage} (a) and (b) shows that SS enables the model to better localize the fine-grained target than RRC. Comparing to Figure~\ref{fig:vis_stage} (a)(b), Figure~\ref{fig:vis_stage} (c) shows that \textit{dual-stage} training further improves localization than the \textit{first-} or \textit{second-stage only}.

\section{Related Work}
\label{sec_related_work}
Self-supervised methods have largely reduced the performance gap between supervised models on various downstream vision tasks. Most early methods design hand-crafted pretext tasks~\cite{iccv2015relativeloc, eccv2016coloring, iclr2018rotation}. For example, \citet{iclr2018rotation} proposed learning image features by training the network to predict different rotation angles of the same image. Doersch\textit{et al.}~\cite{iccv2015relativeloc} proposed learning image features by training the network to predict the correct order of random cut image patches. These pretext tasks rely on somewhat ad-hoc heuristics, which limits the generality of learned representations.

Recently, contrastive learning~\cite{cvpr2020moco, 2020simclr, 2020mocov2, nips2020swav, 2021Barlow, Li2021GenURL, eccv2022dlme} achieved state-of-the-art performance, which learns instance-level discriminative representations by contrasting positive pairs against negative pairs. In particular, MoCo~\cite{cvpr2020moco} adopts a memory bank to store negative samples while embedding different views of the same image using an online encoder and a momentum encoder. SimCLR~\cite{2020simclr} yields comparable results using the sufficiently large batch size in replacement of the memory bank. Getting rid of the notion of negative pairs, BYOL~\cite{nips2020byol} pulls together positive pairs generated by online and momentum encoders with the help of a predictor and the stop gradient mechanism. Another popular form of self-supervised learning is clustering-based methods~\cite{eccv2018deepcluster, cvpr2020odc, nips2020swav}. Without computing pairwise comparisons, SwAV~\cite{ nips2020swav} maps image features to a set of learnable prototype vectors. Various mechanisms are proposed to learn useful representations rather than trivial solutions in contrastive methods such as a constant representation. 

More recently, some research endeavors have been made on top of contrastive methods to enhance pre-training quality for specific downstream tasks, such as object detection and segmentation~\cite{cvpr2021propagate, iccv2021detco, xiao2021region,liu2020self,li2021dense}. Most methods adopts detection or segmentation components to learn pixel-level contrastive representation~\cite{cvpr2021InsLoc, iccv2021MaskContrast, iccv2021updetr, selvaraju2021casting}. DSC~\cite{li2021dense} proposed to model pixel-level semantic structures within images by taking into consideration the semantic relations of both intra- and inter-image pixels. Self-EMD~\cite{liu2020self} learns representations by measuring the similarity among all location pairs using the earth mover's distance (EMD). LooC~\cite{iclr2021looc} proposed to construct separate embedding sub-spaces for each augmentation instead of a single embedding space. DiLo~\cite{aaai2021DiLo} proposed a copy-paste~\cite{ghiasi2021simple,remez2018learning} based augmentation approach that randomly pastes masked foreground onto a variety of backgrounds. CASTing~\cite{selvaraju2021casting} crops views based on a ratio threshold of the area of saliency regions (required mask-level supervision) to the area of the cropped patch. However, existing methods are trained on general-purpose coarse-grained datasets while neglecting fine-grained scenarios where low-level background texture features provide little clue to the category information of the foreground subject. To address this issue, we propose a \textit{dual-stage} pre-training pipeline that utilizes coarse- and fine-grained datasets for better fine-grained representation learning. Appendix~\ref{sec:dis} provides a discussion of the relationship between our proposed CVSA and previous methods.

Current efforts \cite{xiao2015application, simon2015neural, zheng2019looking, huang2020interpretable, ding2019selective, zhao2021graph} in fine-grained recognition are primarily dedicated to fine-tuning model pre-trained on supervised ImageNet either by localizing distinct parts or by learning fine-grained features. However, there exists little exploration in self-supervised pre-training for fine-grained categorization. In the paper, we attempt to bring localization and fine-grained feature representation learning to the pre-training stage, using fine-grained datasets. To address this issue, we propose a \textit{dual-stage} pre-training pipeline that utilizes coarse- and fine-grained datasets for better fine-grained representation learning.

\section{Conclusion}
In this paper, we find that learning to extract discriminative texture information and localization are equally crucial for fine-grained self-supervised pre-training with empirical analysis.
We proposed a \textit{dual-stage} pre-training pipeline with the \textit{first-stage} to train feature extraction and the \textit{second-stage} to train localization. To empower the model with localization abilities in the \textit{second-stage}, we propose \emph{cross-view saliency alignment} (CVSA), a new unsupervised contrastive learning framework. Extensive experiments on fine-grained benchmarks demonstrate the effectiveness of our contributions in learning better fine-grained representations.

\section*{Acknowledgement}
This work is supported by the Science and Technology Innovation 2030- Major Project (No. 2021ZD0150100) and the National Natural Science Foundation of China (No. U21A20427). We thank Zicheng Liu, Yifan Zhao, and all reviewers for polishing the writing.


\bibliography{cvsa}

\clearpage
\renewcommand\thefigure{A\arabic{figure}}
\renewcommand\thetable{A\arabic{table}}
\setcounter{table}{0}
\setcounter{figure}{0}

\appendix
\section{Fine-grained Pre-training Essentials}
\label{sec_empirical_study}
We evaluate the capabilities learned out of three classes of pre-training mechanisms, namely self-supervised contrastive, non-contrastive, and supervised methods. In particular, we focus on discriminative feature extraction and object localization ability. Without loss of generality, we select MoCo.v2, BYOL, Rot-Pred, and supervised classification for comparison. To explore the effects of object localization, we develop a simple binary classification as a pre-training task where the model is asked to classify images from CUB as foreground class and images from COCO as background class. (rather than design detection specific modules as InsLoc~\cite{cvpr2021InsLoc}).

\subsection{Experimental Setup}
\label{sec_settings}
\paragraph{Dataset.} 
We evaluate the performance of baselines' representation pre-trained on the training set of 100$\%$ ImageNet-1k (IN-1k), 10$\%$ IN-1k, COCO, and CUB, details of datasets used are described in Sec.~\ref{sec_exp}. We use the same fixed split for the 10$\%$ IN-1k where we randomly sample 10$\%$ of the total training set size from each class.

\paragraph{Pre-training details.} 
To ensure impartial comparisons, MoCo.v2 data augmentations are adopted for all self-supervised methods and follow the exact setup described in the original papers. OpenMixup~\cite{li2022openmixup} is adopted as the codebase. All models are pre-trained 200 epochs on 100$\%$ IN-1k and 800 epochs on other datasets. For the binary classification, the model is pre-trained by SGD optimizer with an initial learning rate of $0.1$ adjusted by a cosine annealing scheduler, the SGD momentum of $0.9$, and the weight decay of $0.0001$.
In contrastive learning pre-training, the input resolution is $224\times 224$ and the data augmentation strategy follows MoCo.v2~\cite{2020mocov2} as following: Geometric augmentation is \textit{RandomResizedCrop} with the scale in $[0.2,1.0]$ and \textit{RandomHorizontalFlip}. Color augmentation is \textit{ColorJitter} with \{brightness, contrast, saturation, hue\} strength of $\{0.4, 0.4, 0.4, 0.1\}$ and an applying probability of $0.8$, and \textit{RandomGrayscale} with an applying probability of $0.2$. Blurring augmentation is using a square Gaussian kernel of size $23\times 23$ with a standard deviation uniformly sampled in $[0.1, 2.0]$. During the evaluation, images are resized to $256$ pixels along the shorter side and are center cropped to $224\times 224$.

\paragraph{Evaluations.}
We evaluate the learned representation with a linear evaluation protocol, and a fully supervised fine-tune evaluation protocol. The  \textbf{linear evaluation protocol} is a commonly adopted protocol detailed in~\cite{2020simclr}\cite{cvpr2020moco}, i.e., train a linear classifier on top of the frozen representation on the labeled training set. We use SGD optimizer with a cosine annealing scheduler, the SGD momentum of $0.9$, and the weight decay of 0. Based on supervised fine-grained classification settings, we adopt the batch size of 16 with 50 training epochs for small-scale fine-grained datasets, while using the batch size of 32 with 80 training epochs for iNat2018. To avoid evaluation deviations caused by the learning rate, we report the best test performance achieved among the initial learning rate in $\{0.1, 0.01, 0.001\}$ for each comparing method. The fully supervised \textbf{fine-tune evaluation protocol}, as proposed in~\cite{zhai2019s4l, 2020simclr}, fine-tunes the entire network on the training set with labels. Since the original protocols are designed for coarse-grained datasets such as IN-1k, we adopt fine-grained training settings: we use SGD optimizer with a cosine annealing scheduler and a batch size of 16 training 50 epochs. To avoid evaluation deviations, we sweep over the initial learning rate in $\{0.1, 0.05, 0.01, 0.005, 0.001\}$ and the weight decay in $\{0.0005, 0.0001\}$, and select the hyperparameters achieving the best performance on the validation set. The linear test accuracy of the pre-trained model is referred to as the model's discriminative feature extraction ability~\cite{iccv2019scaling}. We refer to the fine-tune evaluation as a pre-training representation quality metric for fine-grained classification problems in a practical sense. For each method, we report mean top-1 accuracy on the test set over 3 trials. To evaluate the localization ability of different approaches, we use a class activation mapping (CAM) based metric \textbf{MaxBoxAcc}~\cite{choe2020cvpr}. A larger MaxBoxAcc indicates better localization ability.

\begin{table}[t]
    \centering
    \setlength{\tabcolsep}{0.8mm}
    \resizebox{1.0\columnwidth}{!}{
    \setlength\tabcolsep{3pt} 
\begin{tabular}{l|ccc|ccc|ccc|ccc}
\toprule
           & \multicolumn{3}{c|}{100\% IN-1k}           & \multicolumn{3}{c|}{10\% IN-1k}            & \multicolumn{3}{c|}{COCO}                  & \multicolumn{3}{c}{CUB}                    \\ \cline{2-13} 
Methods    & Finetune     & Linear       & MaxBoxAcc    & Finetune     & Linear       & MaxBoxAcc    & Finetune     & Linear       & MaxBoxAcc    & Finetune     & Linear       & MaxBoxAcc    \\ \hline
Supervised & \gray{79.25} & \gray{67.64} & \gray{52.18} & \gray{74.68} & \gray{63.07} & \gray{51.42} & \gray{65.41} & \gray{61.39} & \gray{49.87} & \gray{67.82} & \gray{49.85} & \gray{37.31} \\
Rot-Pred   & 67.66        & 25.29        & 46.14        & 68.81        & 23.71        & 47.22        & 67.95        & 16.54        & 43.39        & \bf{66.67}   & \bf{15.46}   & \bf{34.69}   \\
MoCo.v2    & 73.19        & \bf{33.34}   & 47.73        & 69.62        & \bf{25.32}   & \bf{48.23}   & 68.47        & \bf{18.69}   & \bf{46.74}   & 63.21        & 15.03        & 33.36        \\
BYOL       & \bf{76.63}   & 29.17        & \bf{49.23}   & \bf{70.42}   & 20.38        & 46.52        & \bf{68.55}   & 16.36        & 45.60        & 64.85        & 15.24        & 33.81        \\
\bottomrule
\end{tabular}
    }
    \vspace{-0.75em}
    \caption{\textbf{Comparison of pre-training methods.} Top-1 fine-tune and linear accuracy and MaxBocAcc (\%) are reported.}
    \label{tab:table1}
    \vspace{-1.5em}
\end{table}

\subsection{Essential Requirement and Formulation}
\label{sec_formulation}
\paragraph{Where does the gap lie between self-supervised and supervised pre-training?}
As shown in Table~\ref{tab:table1}, when pre-trained on 100$\%$ IN-1k and 10$\%$ IN-1k, the supervised method consistently outperforms all self-supervised pre-training methods. Compared to supervised pre-training, all self-supervised approaches yield lower MaxBoxAcc, indicating a lack of localization ability. Self-supervised methods are task-agnostic and could only learn low-level features, i.e., gradient and direction-dependent features for rotation, and invariant features across views to cluster different objects for contrastive methods. However, the supervised method discards task-irrelevant information and extracts related semantic features. Deep CNN, such as ResNet, has its natural ability in localization during the supervised pre-training process. However, such localization ability could hardly be acquired during self-supervised pre-training. Also, notice that supervised pre-training on CUB yields much lower linear and fine-tune accuracy than self-supervised methods, which states that discriminative feature extraction is largely affected by the size of the dataset.


\begin{wrapfigure}{r}{0.60\textwidth}
    \centering
    \vspace{-1.5em}
    \includegraphics[width=1\linewidth]{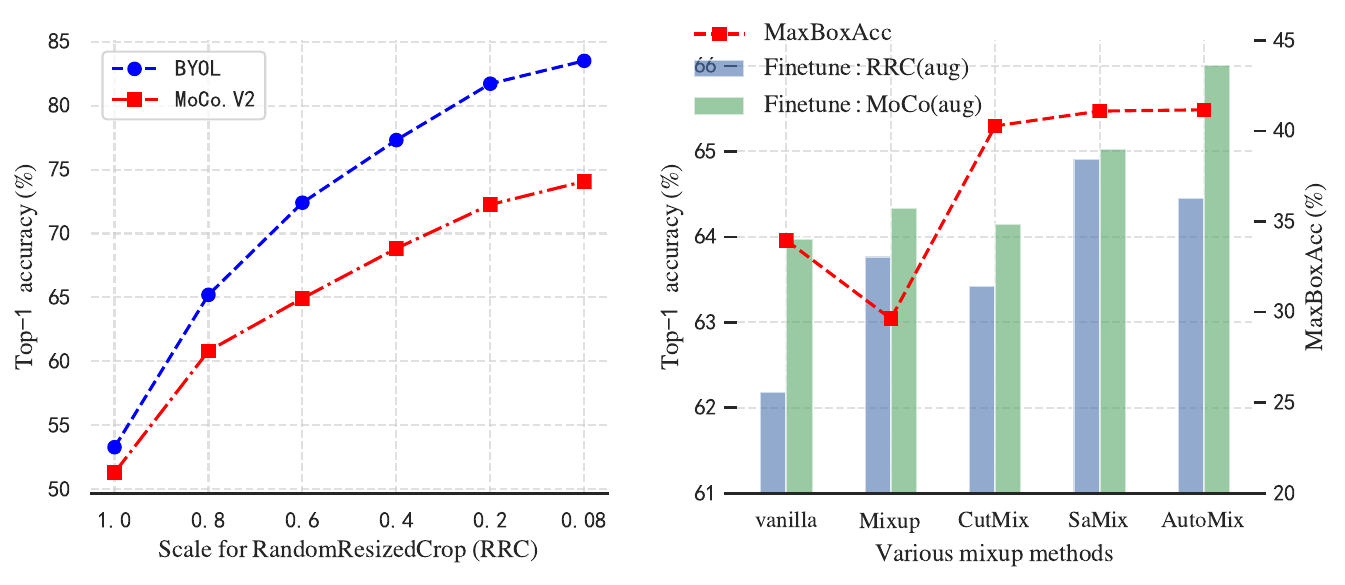}
    \vspace{-2.25em}
    \caption{Left: Performance analysis of \textit{RandomResizedCrop} (RRC) on STL-10 based on BYOL and MoCo.v2. The $x$ axis indicates the scale factor for RRC and the Top-1 accuracy (\%) of fine-tune evaluation is reported. Right: Performance analysis of the binary classification task with various mixup methods. The Top-1 accuracy (\%) of fine-tune evaluation and MaxBoxAcc are reported.
    }
    \vspace{-2em}
    \label{fig:analysis}
\end{wrapfigure}

\paragraph{Why doesn't the contrastive method look at the bird?}
We hypothesize that the lack of localization ability comes from the commonly adopted \textit{RandomResizedCrop} (RRC) augmentation, where a random size patch at a random location is cut from the original image and then resized to the original size. We verify the hypothesis on STL-10~\cite{2011stl10} (a subset of IN-1k) based on BYOL and MoCo.v2 using their training settings on IN-1k. Figure~\ref{fig:analysis} left shows that performances of both methods drop drastically as the cropped patch scale enlarges. The best accuracy is achieved with a scaling factor of 0.08, which is the default hyperparameter choice for current contrastive-based approaches. When asked to pull together two overly small patches cut from the same image, the model is forced to exploit low-level local texture features leading to poor localization ability. 
Different from contrastive-based methods, Rot-Pred takes in whole images rotated by four degrees as input. The authors claim that the model is required to understand the location and pose of the objects depicted in the image in order to predict the rotation angle. As can be seen from Table~\ref{tab:table1}, Rot-Pred indeed yields better localization ability and overall performance than contrastive-based algorithms on CUB.

\begin{wrapfigure}{r}{0.50\textwidth}
    \centering
    \vspace{-2em}
    \includegraphics[width=1\linewidth]{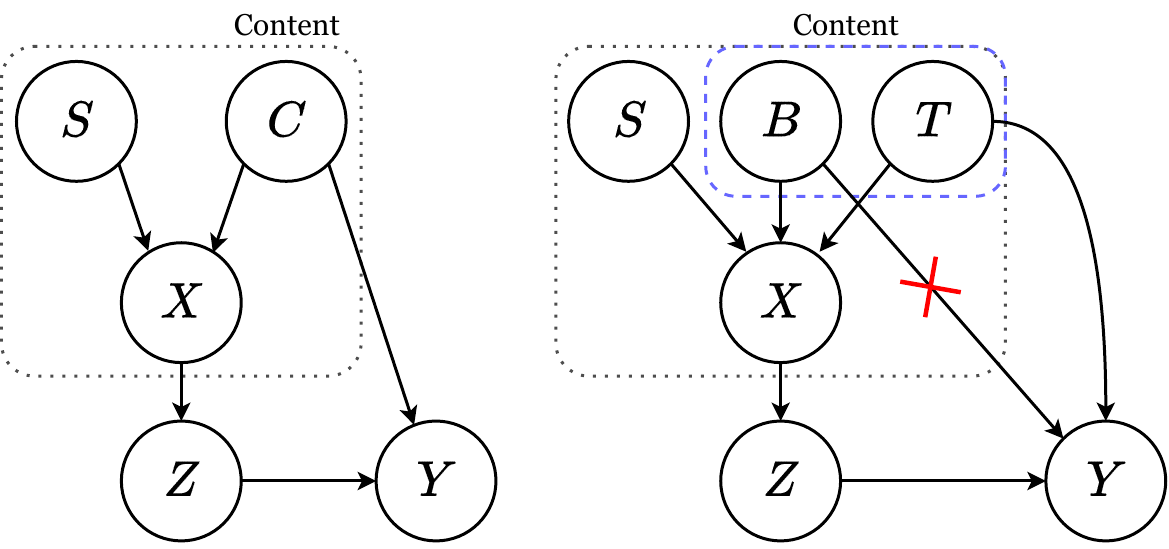}
    \vspace{-2em}
    \caption{Causal interpretation of existing contrastive methods (left) vs.\ CVSA (right) from a causal perspective. The direct link denotes the causality from the cause to the effect.
    }
    \label{fig:causal}
    \vspace{-1em}
\end{wrapfigure}
Then, we formulate current contrastive methods with a causal graph as illustrated in Figure~\ref{fig:causal} and will later use this concept to formalize our pre-training scheme. Let $X$ be images with the content $C$ composed of background prior $B$ and foreground target prior $T$, generated with style prior $S$ as augmentations like color jittering. Latent representation $Z$ is learned and used to infer image labels $Y$. Contrastive methods assume image labels $Y$ are an effect of whole image content $C$ (both $B$ and $T$) due to the local texture-biased nature. In this work, we propose to weaken the causality between $B$ and $Y$ to make $Y$ a more direct effect of $T$.

\paragraph{Is localization all you need?}
We report linear and fine-tune test accuracy and MaxBoxAcc of the binary classification on the CUB test set pre-training with various mixup augmentations~\cite{zhang2017mixup, uddin2020saliencymix, liu2022automix,yun2019cutmix} as well as image augmentations in Figure~\ref{fig:analysis} right. From Table \ref{tab:table1} and Figure~\ref{fig:analysis}, it is observed that the fine-tuned model using binary classification as pre-training yields comparable MaxBoxAcc as supervised pre-training, which indicates that a simple binary classification supervision signal empowers the network with localization ability. Yet, there still exists a vast fine-tuned accuracy gap compared to supervised pre-training on IN-1k. We assume that the gap mainly comes from an inferior feature extraction ability of the binary classification pre-training, as could be conducted from a much lower linear test accuracy. In other words, for better fine-grained recognition pre-training, discriminative feature extraction ability is as essential as localization ability.

Next, we investigate how different mixup augmentations affect the model's localization ability. We notice a simple interpolation between images as done by Mixup negatively impacts the model's localization ability. This negative impact is largely due to the unnatural characteristics of the mixed images. Cutmix mixes samples by replacing the image region with a patch from another training image, while SaliencyMix replaces the image region with the saliency region from another training image. We observe that both CutMix and SaliencyMix bring about better localization ability. However, directly applying these mixup-based augmentations to contrastive learning leads to degenerate solutions. Contrastive learning essentially expects positive pairs to share common semantic objects while keeping negative pairs as much dissimilar as possible. Due to the randomness introduced by such mixup algorithms, augmented images may contain multiple semantic objects or contain no semantic object at all. Without proper supervision, this easily causes the learned representation space to collapse during self-supervised contrastive pre-training. We address this problem by proposing an image augmentation technique that swaps saliency regions of images which aims to introduce solely background variation.

\paragraph{Formulation.}
From the previous analysis, given a fine-grained classification problem, similar to \cite{arora2019theoretical}, we assume $\mathcal{X}$ to be a set of all samples with an underlying set of discrete latent classes $\mathcal{C}$ that represent semantic content, we obtain the joint distribution between each sample $x$ and its class $c$:
\vspace{-0.5em}
\begin{align}
    p(c,x) = p(c|x_{fore}) \cdot p(x_{fore}|x),
\end{align}
where $x_{fore}$ stands for the foreground object. This factorization captures two important intuitions: (1) Given an image of a fine-grained object; the model should first localize the foreground object ($p(x_{fore}|x)$), namely, the localization ability of the model. (2) To further tell the species of the foreground object ($p(c|x_{fore})$), discriminative texture features should be extracted, namely, the texture extraction ability of the model. Following this formulation, a \textit{dual-stage} pre-training pipeline is naturally proposed for self-supervised fine-grained recognition. In particular, we refer to previous contrastive learning methods such as MoCo.v2 and BYOL on large datasets such as ImageNet or COCO as the \textit{first-stage} and the proposed CVSA as the \textit{second-stage}. The model's discriminative texture extraction ability could be fulfilled by \textit{first-stage} pre-training. In the \textit{first-stage}, we regard the image content as a whole as the same assumption of current contrastive methods. For the \textit{second-stage} pre-training, we propose a framework called cross-view saliency alignment (CVSA) to enhance the model's localization capability.

\section{More Ablation Experiments}
\label{app_ablation_detection}
We further study the impact of using saliency information provided by different saliency detection methods based on experiment settings in Sec.~\ref{subsec:ablation}. We compare five well-recognized saliency detection methods (VSFs~\cite{2010VSFs_opencv}, GS~\cite{eccv2012GS}, FST~\cite{cvprw2014AFS}, RBD~\cite{cvpr2014RBD} and BSANet~\cite{cvpr2019BSANet}) and the ground truth bounding box on CUB. As shown in Table~\ref{tab:ablation_saliency}, the proposed SaliencySwap and alignment loss are robust to the quality of saliency bounding boxes because our approach helps the network to localize the object roughly and extract fine-grained semantic features. It is no need to provide accurate segmentation masks of the foreground objects as in object detection and segmentation~\cite{cvpr2021casting,iccv2021MaskContrast}.
\begin{table}[ht]
    \vspace{-0.5em}
    \centering
    \resizebox{0.80\columnwidth}{!}{
    \begin{tabular}{lcccccc}
        \toprule
        Method        & VSFs  & GS    & FST   & RBD   & BSANet     & Groundtruth \\
        \hline
        BYOL+SS       & 64.27 & 64.32 & 64.28 & 64.34 & 64.33      & \bf{64.35}  \\
        BYOL+SS+Align & 64.89 & 64.94 & 64.93 & 64.97 & \bf{65.04} & 65.02  \\
        \bottomrule
    \end{tabular}}
    \vspace{0.25em}
    \label{tab:ablation_saliency}
    \caption{\textbf{Evaluation of different saliency detection methods for \textit{second-stage only} pre-training.} Top-1 accuracy (\%) under fine-tune evaluation is reported on CUB.
    }
    \vspace{-0.5em}
\end{table}

\begin{table}[h]
    \centering
    \vspace{-0.5em}
    \resizebox{0.85\columnwidth}{!}{
    \setlength{\tabcolsep}{1.0mm}
    \begin{tabular}{l|cc|ccc|ccc|ccc}
    \hline
        & Settings & V100      & \multicolumn{3}{c|}{BYOL} & \multicolumn{3}{c|}{BYOL+DiLo}     & \multicolumn{3}{c}{BYOL+\bf{CVSA}}      \\ \hline
CUB-200 & 400 ep   & 1$\times$ & 6.0h   & 30.3M   & 72.5   & \red{+0.5}h & 45.7M & \green{+4.1} & \red{+0.5}h & 45.1M & \bf{\green{+4.6}} \\
NAbirds & 400 ep   & 4$\times$ & 7.5h   & 30.3M   & 76.1   & \red{+2.0}h & 45.7M & \green{+2.9} & \red{+1.5}h & 45.1M & \bf{\green{+3.5}} \\
    \bottomrule
    \end{tabular}}
    \caption{\textbf{Comparison of computational overhead.} The total training time (hours), the number of parameters (M), and the performance of the \textit{dual-stage} setting are reported.
    }
    \label{tab:ablation_time}
    \vspace{-0.5em}
\end{table}

We then compare the computation overhead and the performance gain in Table~\ref{tab:ablation_time}, demonstrating that the proposed CVSA significantly improves BYOL with limited extra computational overhead.

\section{Discussion}
\label{sec:dis}
For fine-grained classification, our proposed CVSA aims to balance the abilities of target localization and discriminative feature extraction. 
As for contrastive-based methods designed for downstream tasks like object detection and segmentation, most frameworks perform object localization and classification in two network branches, focusing on improving localization ability. 
Compared to SaliencyMix~\cite{uddin2020saliencymix} and DiLo~\cite{aaai2021DiLo}, DiLo randomly places the masked foreground objects to raw background images such as texture backgrounds, while our proposed SaliencySwap swaps the saliency region in the randomly selected background image and the source image and regards the augmented view as a positive sample, as shown in Figure~\ref{fig_1}. Notice that SS and SaliencyMix only require coarse saliency information described by bounding boxes, while DiLo uses pixel-wise saliency masks.
Compared to CASTing~\cite{cvpr2021casting}, it improves object localization by cropping views based on saliency regions (required mask-level supervision) and maximizing the similarity between learned saliency masks. However, for fine-grained classification, both discriminative feature extraction and target localization are crucial (performed in the same branch). Our alignment loss utilizes saliency maps of two images and aligns them with cross attention of the two views.


\end{document}